\pdfoutput=1

\documentclass[11pt]{article}

\usepackage[final]{coling}

\usepackage{times}
\usepackage{latexsym}

\usepackage[T1]{fontenc}

\usepackage[utf8]{inputenc}

\usepackage{microtype}

\usepackage{inconsolata}

\usepackage{graphicx}

\title{The Theater Stage as Laboratory:\\Review of Real-Time Comedy LLM Systems for Live Performance}

\author{Piotr Mirowski \\
  Improbotics \\
  London, UK \\
  \texttt{piotr.mirowski@computer.org}
  \\\And
  Boyd Branch \\
  Improbotics \\
  Coventry, UK \\
  \\\And
  Kory Mathewson \\
  Improbotics \\
  Montréal, Canada \\
}

\newcommand{\revise}[1]{{\color{black}#1}}

\begin{document}
\maketitle
\begin{abstract}
In this position paper, we review the eclectic recent history of academic and artistic works involving computational systems for humor generation, and focus specifically on live performance. We make the case that AI comedy should be evaluated in live conditions, in front of audiences sharing either physical or online spaces, and under real-time constraints. We further suggest that improvised comedy is therefore the perfect substrate for deploying and assessing computational humor systems. Using examples of successful AI-infused shows, we demonstrate that live performance raises three sets of challenges for computational humor generation: 1) questions around robotic embodiment, anthropomorphism and competition between humans and machines, 2) questions around comedic timing and the nature of audience interaction, and 3) questions about the human interpretation of seemingly absurd AI-generated humor. \revise{We argue that these questions impact the choice of methodologies for evaluating computational humor, as any such method needs to work around the constraints of live audiences and performance spaces. These interrogations also highlight different types of collaborative relationship of human comedians towards AI tools.} 
\end{abstract}

\section{Introduction}

Attempting to combine humor with machine intelligence is a long-standing subject of scientific enquiry and it is perceived as a fundamental challenge \citep{raskin1979semantic}. \citet{amin2020survey}, \citet{veale2021your} and \citet{sharples2022story} provide authoritative reviews of this nascent field, which can be supplemented with examples of recent works that rely on large language models \citep{winters2018automatic,toplyn2023witscript3,chen2023prompt,jentzsch2023chatgpt,tikhonov2024humor}.

According to several computational humor researchers including \citet{winters2021computers}, \emph{``humans are the only known species that use humor for making others laugh. Furthermore, every known human civilization also has had at least some form of humor for making others laugh''}
\citep{caron2002ethology,gervais2005evolution}. This observation is often extrapolated into the assertion that humor remains an elusive goal for AI (\revise{in the same vein, researchers in computational storytelling have defined} improvisational storytelling \revise{as a \emph{grand challenge} for AI} \citep{martin2016improvisational}). According to recent surveys \citep{mirowski2024robot}, this skeptical view \revise{about AI's comedic potential} is a strongly-held opinion shared by the wider comedy community, from actors and audiences to reviewers and journalists writing about comedy.

For this reason, we posit that comedy audiences and performance spaces are the ultimate environments to critically evaluate the quality of systems for the computational generation of humor. Even though some studies like \citep{gorenz2024funny} have evaluated AI-generated humor using crowd-sourced workers, human-computer interaction researchers have raised concerns about the poor quality of crowd-sourced evaluation of open-ended text generation \citep{karpinska2021perils}, which can be attributed to lack of buy-in from evaluators, and to missing context. The setup of live professional comedy invites {\bf paying audiences} expecting to have a good time, and {\bf critical reviewers} judging the overall performance: it thus provides with a realistic and challenging testbed for computational humor systems. Moreover, the live nature of comedy performance \revise{creates} rich, {\bf interactive exchanges} between comedians and audiences, which---unlike pure online evaluation---allows comedians (and comedy generation systems) to incorporate {\bf real-time feedback} and rich sensory and {\bf cultural context} in the {\bf safe} environment of theater.

At the intersection of live theater and humor sits improvisational theater and comedy \citep{johnstone1979impro}, a complex theatrical art form that can be traced back to Rhapsodes in Ancient Greece or to Commedia dell'Arte \citep{lea1934italian,collins2001improvisation,mathewson2019humour}. Because it relies on natural human interaction and demands constant adaptation to an evolving context, theatrical improvisation (like jazz) has been qualified as \emph{``real-time dynamic problem solving''}~\cite{magerko2009empirical,johnson2002jazz}. According to \citet{mathewson2018improbotics} \emph{``improv requires performers to exhibit acute listening to both verbal and non-verbal suggestions coming from the other improvisers, split-second reaction, rapid empathy towards the other performers and the audience, short- and long-term memory of narrative elements, and practiced storytelling skills''}, making it a highly interesting challenge for AI systems. \revise{Interestingly, theatrical improvisation encourages risk taking and experimentation, and it even ``celebrates failure'' thanks to a tacit agreement between improvisers and audiences who acknowledge the challenge of making up comedic material live on the stage. The improv stage thus provides a ``safe'' environment to test technological tools like artificial intelligence.}

What follows is a literature and performance review of the state of the art of computational humor systems deployed in real-time in front of live audiences, whether in physical or in virtual spaces. We group these according to the type of scientific or artistic questions that they raise, starting with questions around robotic {\bf embodiment} of chatbots, anthropomorphism and {\bf competition} between humans and machines (Section \ref{sec:humanity}), and questions around {\bf liveness}, {\bf timing} and utility in the artistic process (Section \ref{sec:live}). We \revise{address} the human {\bf interpretation} and {\bf justification} of seemingly absurd AI-generated humor (Section \ref{sec:language}) \revise{and finish with a discussion on how the constraints of live audiences and performance spaces impact the choice of methodologies for evaluating computational humor (Section \ref{sec:discuss}). We therefore suggest that the setting of live performance allows to define collaborative relationships between human comedians and AI tools.}

\section{Robot comedy as a test of humanity}
\label{sec:humanity}

As introduced in the previous section, a commonly held belief is that humor is seen as the last frontier of intelligence \citep{winters2021computers}. Robot comedy can then be seen as a challenge to humanity itself\footnote{This prompted comedy critic \citet{logan2023can} to unwittingly center comedy over other art forms: \emph{``unlike music and visual art, comedy can’t be easily reduced to an algorithm.'' (sic)}}.

\subsection{Can robots deliver comedy on stage?}

Robot embodiment presents with unique challenges of audience reception. Some robotics and theater practitioners like Hiroshi Ishiguro and Oriza Hirata took the route of anthropomorphism \cite{pluta2016theater} to make the robot presence as human-like as possible, whereas others like Tom Sgouros built a custom robotic arm\footnote{\url{https://sgouros.com/judy/}} or even, like Annie Dorsen, simply staged two laptops ``talking'' philosophy\footnote{\url{https://anniedorsen.com/projects/hello-hi-there/}}.

In 2010, social roboticist Prof. Heather Knight\footnote{\url{https://www.ted.com/talks/heather_knight_silicon_based_comedy}} pioneered staged comedy performances with an Aldebaran Nao\footnote{\url{https://www.aldebaran.com/en/nao}} robot delivering human-written comedy and gathering audience feedback thanks to camera sensors that track audience sentiment following each line delivered by the robot, and used this information to modify next joke selection based on audience feedback \citep{knight2011savvy}.
Starting in 2014, Austin, Texas-based multidisciplinary artist and engineer Arthur Simone staged toy-like humanoid robots to be his partners in improvised theater performances: \emph{Bot Party: Improv Comedy with Robots}\footnote{\url{https://www.botparty.org/}}, thus investigating how to improvise with a robot.

In 2016, theater improvisers and robotics researchers Dr. Piotr Mirowski and Dr. Kory Mathewson from duo \emph{HumanMachine}\footnote{\url{https://humanmachine.live}} developed large language models \citep{sutskever2014sequence} for improvisational comedy \citep{mathewson2017improvised}. Unlike previous, rule-based AI methods geared at generating comedy, they trained general conversational models \citep{vinyals2015neural} trained on OpenSubtitles \citep{tiedemann2009news}. The language model was coupled with speech recognition to listen to their human partner, text-to-speech and text-dependent robotic control to operate a small scale robot such as the Nao or EZ-Robot JD Humanoid\footnote{\url{https://www.ez-robot.com/}}. Comedy derived from the human actor attempting to justify whatever the robot said.

\revise{Some of those robot performances incorporated implicit audience feedback \citep{knight2011savvy,mathewson2017improvised}, but we hypothesize that audiences may have evaluated the novelty of the premises of those shows in addition to their comedic quality.}

\subsection{Computational humor presented as a competition between humans and machine}

The recent rapid deployment of AI in the creative fields has raised ethical issues around the cannibalization of creative economies \citep{frosio2023generative} and the lack of consent in how training data for AI was obtained \citep{zhong2023copyright}. As a consequence, the public debate around AI is currently driven by the fear of replacement; as illustrated below, performance artists engaging AI ask the question whether AI-generated humor can ever \emph{match} human level.

In 2023, and in the context of public releases of generative AI tools, and of their subsequent short-term impact on creative industries (contributing to the Writers Guild of America (WGA) labour action), Los Angeles-based comedians Allisson Goldberg and Brad Einstein created \emph{Comedians vs. AI: Stage Against the Machine}\footnote{\url{https://www.comediansvsai.com/}}. Their show featured two teams of comedians, one ``human'' relying only on their skills, another one supported by Gen AI software like ChatGPT and DALL-E. The show evaluated AI in an adversarial context, pitting one team against another, and promising the audiences reassurance about limited capabilities of the machines; to quote one comedian, ``We have the benefit of having trauma and life experience to pull from that AI doesn't have integrated yet, and that makes us more dynamic and sensitive and hilarious for now.'' \citep{jamerson2023}.

In that same year of 2023, New York-based Comedy Bytes\footnote{\url{https://www.comedybytes.io/}} refined this concept to focus on improvised \emph{rap battles} and \emph{comedy roasts} between a small cast of human performers, and cartoonish virtual avatars puppeteered by actors or text-to-speech, reading AI-generated jokes \citep{tett2023}.

Other improv performances built around adversarial human-AI relationships include \emph{The AI Improv Show} (2023) by London improv school The Free Association\footnote{\url{https://www.thefreeassociation.co.uk/}} (featuring ChatGPT-generated jokes) and Amsterdam-based Boom Chicago who produced \emph{The Future Is Here... And It Is Slightly Annoying}\footnote{\url{https://boomchicago.nl/shows/the-future-is-here/}} (2019) with improv sketches involving a tele-presence robot on wheels connected to a chatbot developed by Botnik Studios\footnote{\url{https://botnik.org/}}.

\subsection{Can an AI Pass the Comedy Turing Test?}

Building upon the idea of human-AI comparison, improv duo \emph{HumanMachine} adapted in 2017 the Turing test \citep{alan1950turing} and introduced its comedic counterpart \citep{mathewson2017turing}. They assembled in 2018 a large-cast improv troupe, \emph{Improbotics}\footnote{\url{https://improbotics.org}}, featuring human actors, some of whom (called \emph{Cyborgs}) get lines from AI via headphones connected to a portable FM radio receiving lines transmitted from the AI chatbot's text-to-speech. Over hundreds of performances, the troupe has devised diverse short-form and long-form improv games featuring the Cyborg in disclosed or concealed identity. In addition to evaluating audiences' perception towards AI, the troupe evaluates audiences' perception of its language capabilities: they \revise{devised a comedy} Turing test by staging non-Cyborg actors who pretend to be controlled by AI \revise{alongside the Cyborg actors}. \revise{One would expect the comedy Turing test to become harder as LLM technology develops, but the comedians invented ways to ``sound like an AI'' to confuse the audiences, thereby demonstrating the limitations of the Turing test.}

Computational humor researcher Dr. Thomas Winters designed, with comedian Lieven Schiere, a more formal Turing test performed on the stage, and aimed at evaluating advances in large language models for writing comedy ahead of the performance \citep{winters2023torfsbotornot}.


\subsection{Comedic deception of audiences by AI}

The idea of deception has been explored in game contexts beyond the Turing test. In 2023, filmmaker Dr. Manuel Hendry designed a dark comedic installation \emph{The Feeling Machine}\footnote{\url{https://www.hendry.me/}}, where a chatbot-powered, ELIZA-inspired ``psychotherapist'' is embodied by an animated mask: once that ``psychotherapist'' establishes a rapport with an individual spectator \citep{hendry2023you}, the system provocatively shows a deep fake of that spectator making up false memories, to raise questions about misuses of technology. \revise{\emph{The Feeling Machine} targets art museum audiences acquainted with ethical discussions around AI; at the opposite side of the spectrum sits a general audience show made by} TV company Endemol Italy and presented in 2023: \emph{Fake Show: Diffidate delle imitazioni}\footnote{\url{https://www.raiplay.it/programmi/fakeshowdiffidatedelleimitazioni}}, an improvisational game show featuring deep fakes.

Company \emph{Improbotics} adapted that concept in 2024: they comically explore alternative life choices of a consenting audience member, acted out by different improvisers who drive live-generated deep fakes \citep{glennon2024boti}.

\section{Live performance and real-time interaction as a test for generative AI}
\label{sec:live}

The commonality behind the shows presented in Section \ref{sec:humanity} was that they addressed ethical interrogations about the role of AI in comedy. In this section, we review shows that investigate how to effectively deliver computational humor on stage.

\subsection{AI co-creating real-time comedy dialogue}

The development of large language models and conversational AI applications mostly focuses on single-user text-based dialogue. Speech recognition and dialogue systems struggle with Multi-Party Chat (MPC). \citet{branch2024designing} describe how they approached this problem in \emph{Improbotics} performances, where multiple actors interact with an AI \emph{Cyborg} stage partner, \revise{just like in a traditional improv comedy show featuring a large cast in a lively performance}. Instead of simple turn-taking in human-chatbot dialogue, the troupe resorts to human-in-the-loop curation of continuously AI-generated lines, where the most comedic or relevant lines are sent to the Cyborg performer via an earpiece; \revise{introducing a second performer who takes responsibility for selecting the AI-generated line creates a ``writer's room'' setup and introduces two levels of human interpretation of AI-generated material.} In their 2024 performances at the Edinburgh Festival Fringe, the troupe replaced earpieces by augmented-reality glasses, delegating the role of AI line curation to the Cyborg performer, \revise{who would simultaneously read some of the AI-generated lines and try to maintain eye contact with stage partners}.

Timing! The most important rule of comedy is... \emph{Improbotics} needed to design both technology (fast speech recognition and language models, and asynchronous generation) to accelerate the robot's responses \citep{branch2024designing}, and dramaturgy (``slow-burn'' improv scenes relying on non-language communication to fill the time lags) \citep{mathewson2018improbotics}. \revise{Improviser Cyborgs expressed they had struggled with AI generated lines because of slow timing and delays; their perception was that the audience preferred timely responses to higher quality but delayed responses.}

In Oregon, Prof. Naomi Fitter \revise{focused on comedy timing as she} has been running since 2019 regular comedy nights where her robot comedian Jon relies on audience laughter to control the timing and delivery of jokes \citep{srivastava2021robot}.

\subsection{AI for inspiration and world building}

Liveness in AI improv shows is not limited to dialogue: human actors can leverage AI-generated ideas for real-time performance. Notably, San Francisco-based Alexa Improvise\footnote{\url{https://ai.nickradford.dev/}} has used an AI assistant for game ideas since 2018; \emph{Yes, Android} by Toronto company \emph{Bad Dog}\footnote{\url{https://baddogtheatre.com/}} featured actors reading LLM-generated lines in 2017; Nouméa-based \emph{La Claque}\footnote{\url{https://laclaqueimpro.com/}} incorporated French-language AI for short-form improv suggestions in 2023; and India-UK troupe \emph{ClimateProv} leveraged Gen AI to inspire climate-themed improvisation in 2022. \citet{winters2019automatically} designed automatic slide generators\footnote{\url{https://talkgenerator.com/}} for \emph{Powerpoint karaoke} games.

Several projects explored LLMs for long-form improvisation by supporting storytelling. Notably \emph{Plays by Bots}, staged since 2022 by Edmonton-based \emph{Rapid Fire Theatre}\footnote{\url{https://rapidfiretheatre.com/}}, rely on scripts co-written with \emph{Dramatron} \citep{mirowski2023co} to build the world for improvisers; and in 2021, \emph{Improbotics} used an AI as narrator for long-form scenes \citep{branch2021collaborative}.

Finally, and while they do not use AI in real time during their performance, many comedians have presented material co-written with AI in front of live audiences, including Darren Walsh\footnote{\url{https://darrenwalsh.co.uk/}} in 2023 and Anesti Danelis\footnote{\url{https://www.anestidanelis.com/}} in 2024 at Edinburgh Fringe.

\revise{The commonality between all those shows is to employ computational humor systems as mere writing tools to support live human performance; as a consequence, audience evaluation is focused primarily on the human performers and how they engage with their audiences.}

\subsection{Live performance with AI in digital spaces}

The development of computer-mediated communication technology has introduced a new way for humans to congregate \revise{and redefined the notion of liveness and audience interaction}. Live performance no longer requires a physical space, as performers and audience can congregate virtually via teleconference and chat, overcoming long geographical distances, as proved by \emph{Failed to Render}\footnote{\url{https://failedtorender.com/}}, a comedy club in virtual reality, or most improv teams performing and rehearsing online during Covid-19.

\citet{branch2023mirror} analysed how shared VR environments and telepresence enhance improvisational flow more than traditional teleconference; a tele-immersive environment was used in 2020-2021 for VR rehearsals and performances\footnote{\url{https://www.art-ai.io/programme/improbotics/}} of \emph{Improbotics}, where the AI agent was represented by an avatar \citep{branch2021tele}. \citet{jacob2019affordance} used computer vision models for physical improv games in \emph{Robot Improv Circus VR}\footnote{\url{https://gvu.gatech.edu/research/projects/robot-improv-circus-vr-installation}}. \emph{PORTAGING} was a humorous prompt battle with Gen AI performed on a Discord channel at NeurIPS 2022\footnote{\url{https://neurips.cc/virtual/2022/56220}}. \revise{In these shows, audience engagement could be measured in chat interactions during streaming and, in some cases, laughter on live audio channels.}

\section{AI language and human interpretation}
\label{sec:language}

The remaining question about computational humor systems for live performance is how they help communicate, or how they challenge human actors to make sense of AI-generated output. 

On one hand, AI can be used for meaning making: multilingual improv in \emph{Rosetta Code} is mediated by speech recognition, machine translation, and in-ear text-to-speech \citep{mirowski2020rosetta}. \revise{Incidently, these three tools are applications that underlie the development of language models}.

On the other hand, we alluded in Section \ref{sec:humanity} to human actors trying to justify ``seemingly absurd'' AI-generated text. Improvisers can leverage LLMs as a creative and acting challenge \citep{mathewson2018improbotics}, and \emph{THEaiTRE}'s scripted production of \emph{AI: When a Robot Writes a Play} exemplifies the glitch aesthetic of involuntarily funny absurdist LLMs \citep{rosa2021robot}. \revise{Absurdist theatre, however, requires supportive audiences.} The \emph{Dramatron} system \citep{mirowski2023co} \revise{was an attempt at making AI-generated theatrical scripts sound less ``absurdist'', and it} aimed at supporting actors by generating more coherent narratives.

\emph{More than Human}, produced in 2019 by Dr. Gunter Lösel, went in the opposite direction. Its human cast (one of whom was taking lines from an LLM) did not attempt to justify those AI-generated suggestions at all. Instead, and following the principles of Dadaism, they used AI to explore and celebrate their own ``inner machine'' \citep{loesel2020digital,losel2024theatre}.


\section{Discussion: evaluation of live AI humor}
\label{sec:discuss}

\revise{This position paper claims that audience feedback from live performances enables a challenging testbed for computational humor systems. Arguably, some comedy material is not amenable to live or improvised formats (e.g., memes, comedic videos and films) as they are pre-written and with no live audience interaction. Nevertheless, these can be assessed by measuring audience engagement on social media, in ratings or at the box office.}

\revise{Human-Computer Interaction literature provides many toolboxes for assessing live audience engagement and the creative process. \citet{branch2024designing} and \citet{mathewson2018improbotics} rely on audience and performer surveys after performances. \citet{srivastava2021robot} measure audience laughter and engagement using microphones. \citet{mirowski2024robot} proposed focus groups with comedians engaging in writing tasks with LLMs and assessing AI using Creativity Support Tool metrics like \citep{cherry2014quantifying,chakrabarty2024creativity}: these metrics can be applied to live and improvisational contexts as well.}

\revise{The fundamental advantage of framing the evaluation of computational humor in the wider context of audience reception and feedback, is that it simultaneously defines the role that AI tools can play in the wider comedy ecosystem--as creativity support tools. Such a framing encourages a collaborative relationship between human comedians and AI tools instead of an adversarial one, and helps approach the various ethical questions around AI art (and comedy in particular) on artists (and comedians) \citep{epstein2023art,jiang2023ai}.}


Humans have used the technologies of their time to tell stories, from cave paintings to the internet. Generative AI is one such technology, and this paper gave examples of storytellers trying to adopt it as a writing tool for performance. Humor and comedy writers can evaluate those tools through real-time human feedback, which can be uniquely provided by live theater---an ideal experimental substrate for creative technology for storytelling.

\section*{\revise{Acknowledgments}}

\revise{The authors wish to thank three anonymous reviewers for instrumental feedback that helped improved the paper, and the casts and guest players of Improbotics for transforming improv and the theatre stage into a laboratory.}

\bibliography{comedy_stage_as_llm_laboratory}




\end{document}